\documentclass[10pt,twocolumn,letterpaper]{article}

\usepackage{3dv}
\usepackage{times}
\usepackage{epsfig}
\usepackage{graphicx}
\usepackage{amsmath}
\usepackage{amssymb}
\usepackage{multirow}

\usepackage{pifont}
\newcommand{\cmark}{\ding{51}}
\newcommand{\xmark}{\ding{55}}

\usepackage[pagebackref=true,breaklinks=true,letterpaper=true,colorlinks,bookmarks=false]{hyperref}

\threedvfinalcopy 

\def\threedvPaperID{118} 

\ifthreedvfinal\fi
\begin{document}

\title{Self-Supervised 2D Image to 3D Shape Translation with Disentangled Representations}

\newcommand{\aand}{\hspace{10mm}}
\author{Berk Kaya \aand Radu Timofte\\
Computer Vision Lab, ETH Z\"urich \\
{\tt\small \{bekaya, radu.timofte\}@vision.ee.ethz.ch }
}

\maketitle

\begin{abstract}
We present a framework to translate between 2D image views and 3D object shapes. Recent progress in deep learning enabled us to learn structure-aware representations from a scene. However, the existing literature assumes that pairs of images and 3D shapes are available for training in full supervision. In this paper, we propose SIST, a Self-supervised Image to Shape Translation framework that fulfills three tasks: (i) reconstructing the 3D shape from a single image; (ii) learning disentangled representations for shape, appearance and viewpoint; and (iii) generating a realistic RGB image from these independent factors.
In contrast to the existing approaches, our method does not require image-shape pairs for training. Instead, it uses unpaired image and shape datasets from the same object class and jointly trains image generator and shape reconstruction networks. Our translation method achieves promising results, comparable in quantitative and qualitative terms to the state-of-the-art achieved by fully-supervised methods\footnote{Our codes and models are available at \url{https://github.com/berk95kaya/SIST.git}.}. 
\end{abstract}

\section{Introduction}

Learning translations between 3D objects and RGB images has become an interesting field in computer vision. Understanding the 3D nature of an object from a 2D view is an ill-posed problem because a single 2D view (or RGB image) may correspond to an infinite number of potential 3D shapes. Therefore, traditional methods usually fail to reconstruct the 3D shape obtained from a single view. On the other hand, deep neural networks can exploit the shape priors provided in the training~\cite{han2019image} and provide plausible 3D reconstructions. 

With the growing interest in single image shape reconstruction tasks, many learning-based approaches were presented~\cite{wu2017marrnet,groueix2018atlasnet,wu2016learning,tatarchenko2017octree,wu2018learning,tulsiani2017multi}. In the past, such models were trained with image-shape pairs provided by datasets such as IKEA~\cite{lim2013parsing} and PASCAL3D+~\cite{xiang2014beyond}. The drawback of these datasets is that they contain only a few samples for each object category due to the difficulties in obtaining annotated shape data.
For this reason, researchers focused on using synthetic datasets such as ShapeNet~\cite{chang2015shapenet} containing textured CAD models. To form image-shape pairs, they used the renderings obtained from corresponding CAD models. This brings the advantage of obtaining an unlimited number of RGB images for each shape. However, the networks trained using this setting often experience a performance drop on camera-captured images because the network cannot adapt itself to a new domain~\cite{pinheiro2019domain}. Relying on direct supervision in single-view 3D reconstruction remains a limitation and self-supervised methods are a promising avenue to explore.

\begin{figure}[t]
\centering
\includegraphics[width=0.43\textwidth]{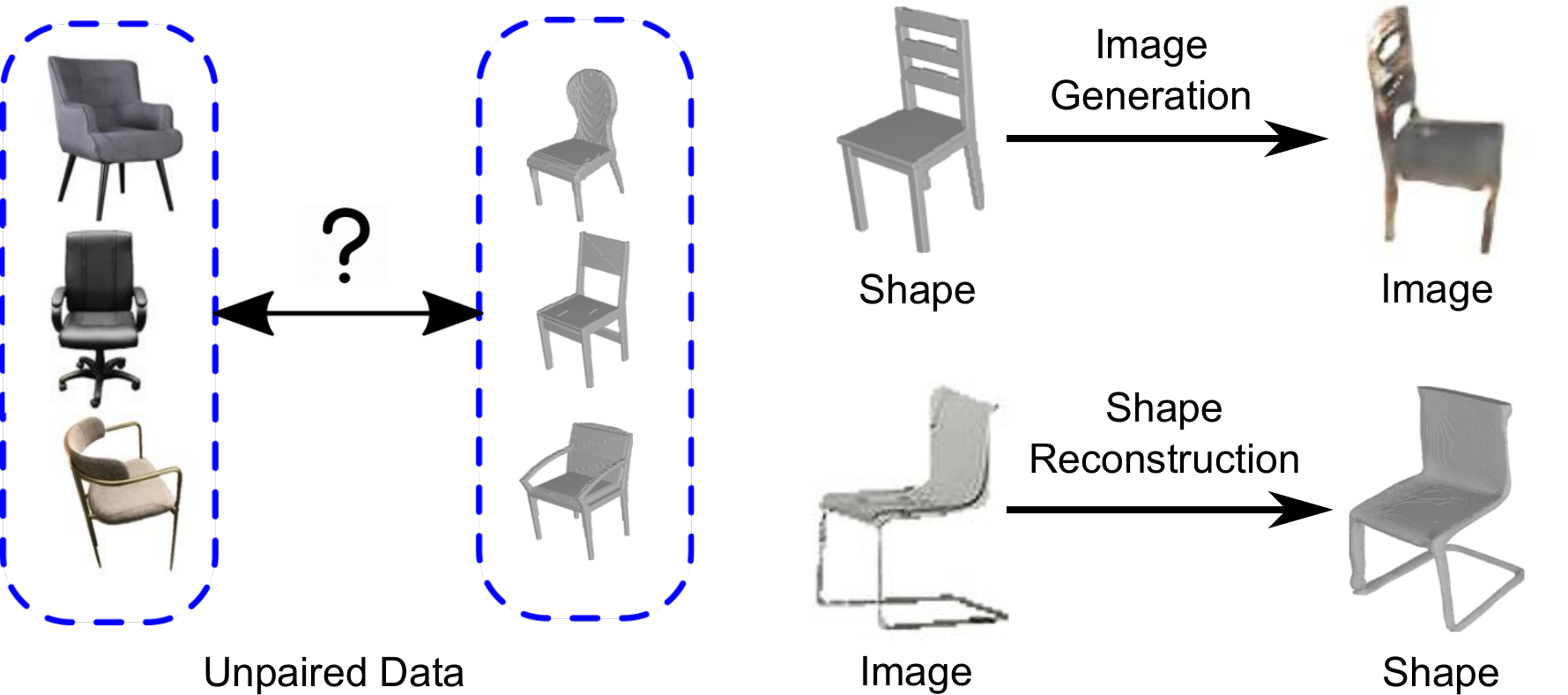}
\caption{ Self-supervised Image to Shape Translation (SIST): Given unpaired image and shape datasets from the same object class, our framework trains image generation and shape reconstruction networks to translate between image and shape domains. }
\label{fig:teaser}
\end{figure}

 Generating realistic images from 3D models is also another open problem in computer vision. 3D shape models do not necessarily contain information of color, texture and reflectance characteristics of the surface. Some datasets contain texture files which can be mounted to triangular mesh structures. Still, reflection/shading algorithms must be applied for rendering RGB images which reflect the shape identity of the object. However, the rendered images often turn out to be non-realistic which causes a domain shift problem. 
 With the recent advances in generative image models, it is possible to generate RGB images with high resolution and visual quality. Generative adversarial networks (GANs)~\cite{goodfellow2014generative} have achieved remarkable results in generating photorealistic images. However, GANs are still restricted in practical usage, since training them is quite difficult and they do not offer a strong control over the generated samples.
Because GANs make use of 2D convolutional operations for image synthesis, they generally tend to ignore the 3D-structure of the generated samples. Therefore, the networks which try to consider the 3D-structure generate blurry images especially in the task of view synthesis.

Motivated by the current limitations in 3D reconstruction and image generation tasks, we propose a novel pipeline that learns disentangled representations and translations between 2D images and 3D shapes from unpaired data. 
Our architecture (described in Fig.~\ref{fig:fullmodel}) uses a renderer to create the depth map of the object shape from a specified viewpoint. Then, it synthesizes an RGB image by sampling an appearance vector from a distribution. Adversarial training is used in this stage to obtain realistic images. Then, we introduce cyclic loss terms to learn the appearance code and viewpoint from the generated image. At the same time, we also introduce a shape reconstruction network to obtain 3D shapes.
The novel combination of these components enables us to obtain translations between image and shape domains. Contrary to other methods in the reconstruction task, \textit{our method does not require paired data}. Our method uses unpaired image and shape datasets from the same class and trains translation functions between these distinct visual domains in self-supervised training. 

Our main contributions are as follows:

\begin{itemize}
\setlength\topsep{0em}
\setlength\itemsep{0em}
\item We propose SIST, a framework that learns translations between RGB images and 3D shapes from unpaired examples.
\item We reconstruct shapes from a single image with an unknown pose.
We use voxel and implicit field~\cite{chen2019learning} representations to model reconstructed 3D objects.
\item We learn disentangled representations of shape, appearance, and viewpoint from an example RGB image. With them, we control several operations such as shape reconstruction, image generation, novel view synthesis, shape\&texture editing and shape\&texture transfer.
\item We also show the benefit achieved by using a small amount of paired examples in a weakly-supervised setting for single-view 3D reconstruction.
\end{itemize}

\section{Related Work}
\subsection{3D Shape Representation}
\textit{Voxels} are the most common and the simplest way to represent 3D shapes. However, due to the caused computational load, other representations were proposed. 
For example, Fan~et~al.~\cite{fan2017point} used a point set generating network to reconstruct a shape and used several postprocessing steps to convert point set to voxel spaces. Similarly, Mandikal~et~al.~\cite{mandikal2019dense} used a hierarchical model to generate a point set prediction of an input image by using three stages. In recent years, mesh-based methods are proposed which perform reconstruction by warping a simple shape~\cite{wang2018pixel2mesh,sahasrabudhe2019lifting,kanazawa2018learning}. AtlasNet~\cite{groueix2018atlasnet} obtained impressive results in mesh-based object generation by warping and combining primitive surface patches.
Recently, learning \textit{implicit fields} to represent shapes has gained popularity. 
Recent works~\cite{mescheder2019occupancy,park2019deepsdf,chen2019learning,saito2019pifu,michalkiewicz2019deep,xu2019disn} have proven that implicit functions achieve superior performance compared to other representations for shape encoding and visual quality.

\subsection{Single View 3D Reconstruction}

Learning-based single view reconstruction methods are usually tackled by encoder-decoder structures. First, the input image is encoded to a latent representation. Then, this representation is used to reconstruct the shape depending on the representation type~\cite{han2019image,tatarchenko2019single}.

Although there are many papers that propose supervised training for 3D reconstruction from images, there are only a few works tackling unsupervised training. 
These methods usually recover shape from a collection of images using a differential renderer~\cite{mees2019self,insafutdinov2018unsupervised,szabo2019unsupervised, kato2019self, henderson2020leveraging}.
Rezende~et~al.~\cite{rezende2016unsupervised} used probabilistic inference networks to train shape generator networks from 2D images without 3D labels. 
Similarly, Yan~et~al.~\cite{yan2016perspective} proposed Perspective Transformer Nets to reconstruct the shape from images.
Henderson and Ferrari~\cite{henderson2018learning} also designed a generative model to learn 3D meshes using rendered images. 
Recently, Gwak~et~al.~\cite{gwak2017weakly} proposed a framework for weakly supervised 3D reconstruction from unpaired data. However, their method requires camera parameters to calculate reprojection errors in raytrace-pooling layer. 
To the best of our knowledge, there isn't any method that effectively utilizes unpaired image and shape datasets for the reconstruction task without additional camera pose information.

\begin{figure*}[t]
\centering
\includegraphics[width=0.84\textwidth]{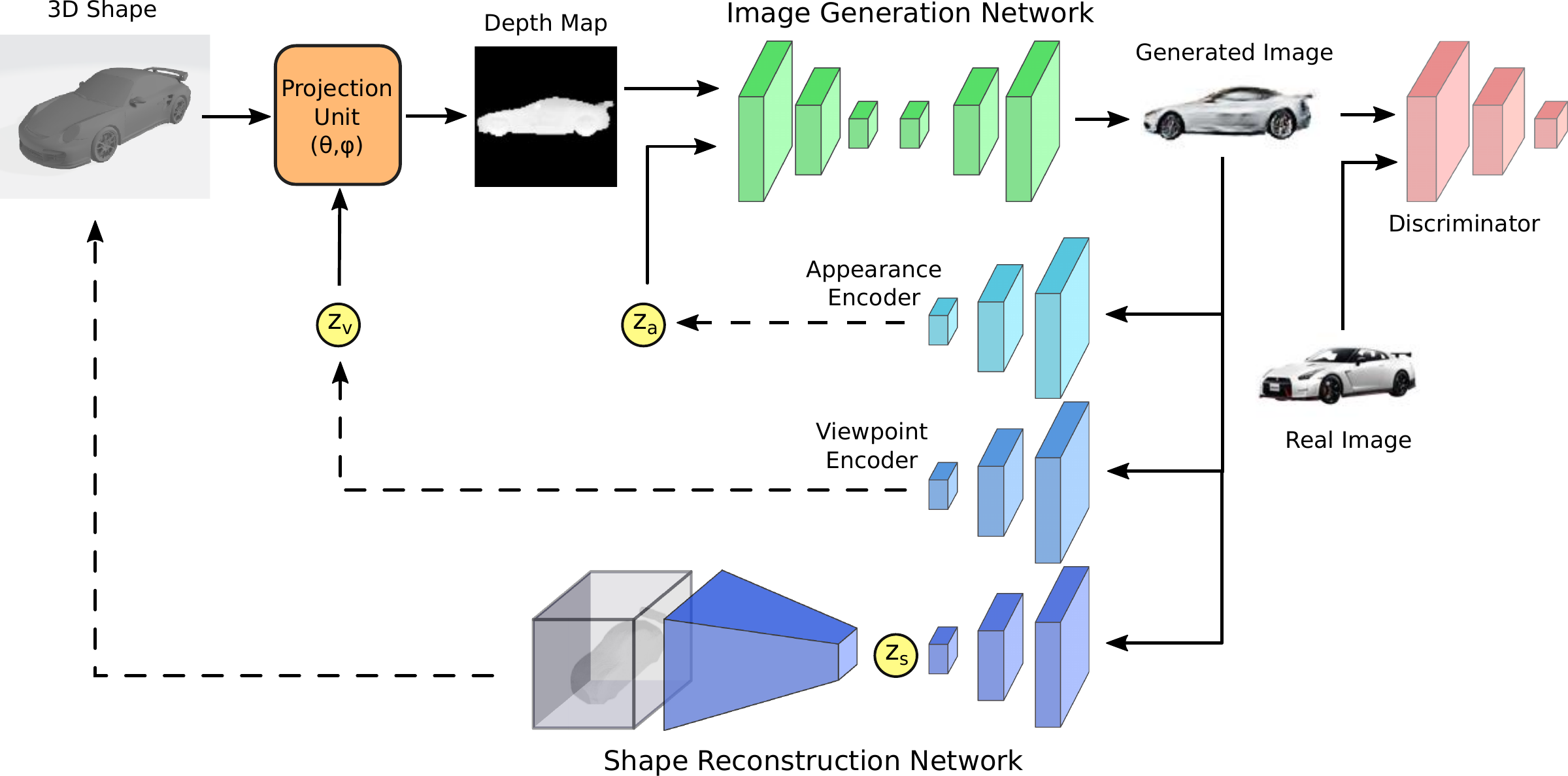}
\caption{\textbf{SIST Architecture:} During training, we generate an RGB image from a shape model by sampling a viewpoint ($z_v$) and an appearance code ($z_a$). The discriminator learns to distinguish between real and generated images. The generated image is also used to train appearance encoder, viewpoint encoder and shape reconstruction network. These modules try to predict the independent components which generate the given image (dashed arrows). }
\label{fig:fullmodel}
\end{figure*}

\subsection{Generative Models}
Apart from 3D reconstruction, our method also generates natural 2D images from 3D shapes. Generative Adversarial Networks (GANs)~\cite{goodfellow2014generative} learn to map samples from a latent distribution to a sample that is indistinguishable from the real data. 
GANs have achieved extraordinary results in many vision tasks such as conditional image synthesis~\cite{gauthier2014conditional,park2017transformation}, image translation~\cite{isola2017image,ignatov2017dslr,romero2018smit}, representation learning~\cite{radford2015unsupervised}, super-resolution~\cite{ledig2017photo,buhler2020deepsee} and domain adaptation~\cite{tzeng2017adversarial}.
Still, only a reduced portion of the proposed works successfully utilize GANs for learning independent representations and controlling them to synthesize images. There are autoencoder based methods for disentangled representation learning~\cite{higgins2016beta,kim2018disentangling,makhzani2015adversarial,hu2018disentangling} but they do not offer explicit control over the factors of interest. They also do not ensure that the learned factors are visually meaningful.

\subsection{Disentangled Image Generation from Shapes}

There are several studies which disentangle pose, shape and appearance-related features. For example, Visual Object Networks (VON)~\cite{zhu2018visual} trains two generative models, one for object shapes and one for images. By combining these two generators with a differentiable renderer, VON controls the shape, texture and pose of the generated object images independently. 
Similarly, HoloGAN~\cite{nguyen2019hologan} proposes a generative model with viewpoint control property. However, it does not explicitly generate a 3D shape to increase the visual quality of the generated samples. Both VON and HoloGAN focus on image generation and their frameworks cannot reconstruct 3D shapes from an image, which our method can.

\section{Proposed Method}

Given unpaired image and shape datasets, our goal is to train networks that map a sample from one domain to another. For this purpose, we introduce two main branches for our network architecture: image generation and shape reconstruction networks. Figure~\ref{fig:fullmodel} illustrates the overall framework. In order to train our networks, we make use of a cycle-consistency loss inspired by CycleGAN~\cite{zhu2017unpaired}. The training begins by picking a voxelized shape representation from our dataset. A renderer is used to create a depth map corresponding to that shape from an arbitrary viewpoint. Then, an encoder-decoder architecture is used to synthesize an RGB image. An appearance vector is introduced in this stage to separate geometric constraints from appearance-based features as in \cite{zhu2018visual,miyauchi2018shape}. We utilize adversarial training to obtain realistic samples. After a fake image is created, we aim to get the shape back with our reconstruction network.  

Different from other methods, our framework is designed to learn many shape-appearance based operations in a single training loop using unpaired texture-less 3D models and RGB images. Assuming these datasets are from the same object category, our method performs translations between 3D and 2D domains. Translating images to 3D shapes stands for single view reconstruction. The inverse of this operation is shape-dependent image synthesis where the generation process is conditioned on a viewpoint and an appearance code. In addition to the translation operations, we also introduce appearance and viewpoint encoders to extract these features from RGB images.

\subsection{Image Generation Network}

\paragraph*{Projection.}
The aim of the projection unit $\mathcal{P}$ is to obtain depth from an input shape. 
In our setting, we use a pin-hole camera model where the camera direction is the negative z-axis. We only define the position of the camera whose axis is aligned with the object center. We also assume fixed camera distance in our projection module and calculate camera calibration matrices using two parameters: rotation around the y-axis (azimuth: $\theta$) and rotation around the x-axis (elevation: $\phi$).  In training, we sample from $\theta \in (-\pi,\pi)$ and $\phi \in (0,\pi/2)$ uniformly such that the rendered images follow a similar distribution with the RGB dataset. The projection unit is the first stage in the training process and it does not have any trainable parameters.

\paragraph*{Shape-Conditioned Image Generation.} 
After obtaining depth maps by sampling a 3D object (shape) and a viewpoint vector, we wish to generate a realistic RGB image that follows the same distribution with our RGB image dataset. For this reason, we train a generator network $G_I$ which takes a depth map and sampled appearance code $z_a$ and creates an RGB image. This problem can be considered as an image translation problem where an input image in the depth domain is translated into RGB by exploiting the surface texture information provided. We handle this task by using adversarial training. Therefore, we also introduce a PatchGAN~\cite{isola2017image} discriminator $D_I$ which tries to discriminate real images from the generated ones. We train our image generation network with Least Squares GAN~\cite{mao2017least} formulation using the following loss function:


\begin{equation}
\begin{aligned}
  \mathcal{L}_I  & = \Big( ~ \mathbb{E}_{x} [(D_I(x))^2]  \\
                 & +\mathbb{E}_{(y,z_v, z_a) } [(1-  D_I(G_I(\mathcal{P}(y,z_v), z_a)))^2] ~ \Big).
  \end{aligned}
\end{equation}

where $x\in X$ represents the real samples from the RGB dataset. $\mathcal{P}(y,z_v)$ is the depth map obtained with the projection unit $\mathcal{P}$ from the shape $y\in Y$ and viewpoint $z_v$.

\subsection{Shape Reconstruction Network}

The next step is to reconstruct a shape given an RGB image. For this task, we propose two design choices, depending on the output shape representation. In both cases, we use the same encoder network architecture $E_S$ which produces the latent vector $z_s$. 
We employ variational training to train a generative shape decoder which produces an object shape from a Gaussian distribution.

\paragraph*{Voxel Decoder.}
Voxel decoder network generates a voxel occupancy grid from a latent representation. In order to generate a 3D occupancy grid from a single input vector, we make use of transposed convolution operations as in 3D-GAN~\cite{wu2016learning}. 
Our architecture generates shapes with a resolution of $128^3$. 
We train the decoder inspired by the cyclic loss. Since we generate an image using a shape from our dataset, our shape reconstruction network simply tries to reconstruct it back.
The reconstructed shape can be expressed with the following formulation:
\begin{equation}
\label{eq:shape}
\hat{y} =   D_S (  E_S (  G_I( \mathcal{P}(y,z_v), z_a) ) )
\end{equation}

where $D_S$ is the shape decoder and $E_S$ is the shape encoder. We pick shape $y$ from our dataset and we sample viewpoint and appearance codes $(z_v, z_a)$. In this way, the error on $\hat{y}$ can be backpropagated through the network. We call this error term $\mathcal{L}_S$ and use mean binary cross-entropy to calculate it. Although we consider all points for voxels, we follow a surface sampling strategy for implicit fields.

\paragraph*{Implicit Field Decoder.}
An implicit field is a continuous function in 3D space. The surface of a shape is described by the level set of the field.
Assuming we have closed shapes, we define the groundtruth implicit function as follows:

\begin{equation}
  \mathcal{F}(u) = \begin{cases} 
      0 & \text{if~point~$u$~is~outside~the~shape} \\
      1 & \text{otherwise}
   \end{cases}
\end{equation}

Generating such a field may be considered as a binary classification problem and we use a multi-layer perceptron (MLP). MLPs can approximate such fields depending on the number of hidden units/layers. By setting a proper threshold for the field ($0.5$ in our case), the surface can be recovered for an arbitrary resolution.

We use the IM-NET decoder architecture proposed by Chen~et~al.~\cite{chen2019learning}. The network uses a feature vector obtained by our encoder architecture and point coordinates as input. It uses the combined vector to estimate the field value of that particular point in the space. To train the network, we sample $K$ points from the voxel space. We sample half of these points from the surface of the object. For the remaining half, we randomly sample such that the number of positive and negative samples are equal. 
Then, we calculate the mean of binary cross-entropy values using these sampled points for the shape loss $\mathcal{L}_S$.

The spatial sampling strategy is only required for training the network. At test time, an occupancy grid can be obtained by applying forward-pass for all points from the defined space. In the end, we again obtain a voxelized shape representation.

\subsection{Learning Appearance and Viewpoint}

Our next task is to learn representations for appearance and viewpoint from the image. For this task, we introduce two encoders which recover appearance and viewpoint from a generated image. The whole image generation process is a translation problem where a depth map, an appearance code, and a viewpoint are mapped to an RGB image. Therefore, we introduce a cyclic loss term inspired by CycleGAN~\cite{zhu2017unpaired} to train appearance and viewpoint encoders $E_A$ and $E_V$.

\begin{equation}
\label{eq:cyclic}
\begin{aligned}
  \mathcal{L}_C =  & \mathbb{E}_{(y,z_v, z_a)} \Big[  \lambda_A\left\|   E_A (G_I( \mathcal{P}(y,z_v), z_a)) - z_a \right\|_1 \\
   +   & \lambda_V \left\| E_V (G_I( \mathcal{P}(y,z_v), z_a)) - z_v \right\|_1 \Big].
   \end{aligned}
\end{equation}

Introducing this term has two advantages. First, it enables us to learn disentangled features from an example RGB image. Second, it prevents the image generator from ignoring the depth map and appearance information.

\subsection{Training Objective}

Our method is the combination of the following blocks: projection unit, image generation network, shape \& appearance \& viewpoint encoders and a shape decoder (voxel or implicit field decoder depending on the design choice). 
The training objective is as follows:

\begin{equation}
\label{eq:full}
  \mathcal{L} =  \lambda_I\mathcal{L}_I +\lambda_S\mathcal{L}_S + \mathcal{L}_C +  \mathcal{L}_{KL}
\end{equation}

We also introduce a KL loss term in order to ensure that shape and appearance representations follow distributions $p(z_s)$ and $p(z_a)$ found in training data. This term pushes the learned representations to a Gaussian distribution so that we can sample these representations to generate new examples.

\begin{equation}
\begin{aligned}
 \mathcal{L}_{KL}= & \mathbb{E}_{(y,z_v, z_a)} \Big[   \lambda_S^{KL} \mathcal{D}_{KL}\big(E_S (G_I(\mathcal{P}(y,z_v), z_a))|| p(z_s)\big) \\
 +  & \lambda_A^{KL} \mathcal{D}_{KL}\big(E_A (G_I(\mathcal{P}(y,z_v), z_a))|| p(z_a)\big) \Big].
 \end{aligned}
\end{equation}

\section{Experimental Results}
\label{sec:experiments}

\subsection{Implementation Details}

In order to demonstrate the performance of our method, we perform experiments on ShapeNet~\cite{chang2015shapenet} and Pix3D~\cite{sun2018pix3d} shape repositories. These datasets provide object-centered voxel data with $128^3$ resolution. For RGB data, we use clean background images with $128\times128$ resolution. We also utilize the dataset provided by VON~\cite{zhu2018visual} which contains unpaired shapes and images for car and chair categories. For more details regarding the datasets, we invite the reader to read the supplementary material.

We assume appearance and shape priors follow a zero-mean unit-variance Gaussian distribution and use this distribution in the calculation of KL loss. We set appearance code length $|z_a|=16$ and shape code length $|z_s|=128$. We use Adam optimizer~\cite{kingma2014adam} with an initial learning rate of $0.0001$ and exponentially decay it after each epoch with a rate of $0.98$. We use a batch size of 16 due to memory restrictions. For the implicit field decoder, we sampled $K=1000$ points in each iteration to train the network. We also used the following hyperparameters in training: 
$\lambda_I = 0.005$,
$\lambda_S = 100$,
$\lambda_V = \lambda_A = 10$, and
$\lambda_S^{KL} = \lambda_A^{KL} = 0.001$ 
We also applied label flipping operation with a probability of $p = 0.05$ in order to prevent the image discriminator from getting too strong against the generator.

\subsection{3D Shape Reconstruction Results}

We start by evaluating the performance of our method on single view shape reconstruction task. In contrast to other methods, we do not rely on supervision. Nevertheless, we compare our performance with the supervised methods to show that we can bridge the gap between supervised and self-supervised methods.

\subsubsection{Comparison with State-of-the-Art} 

To compare our reconstruction results with other works, we perform evaluations on Pix3D dataset~\cite{sun2018pix3d}. Therefore, we collect evaluation results presented in~\cite{wu2018learning} and~\cite{pinheiro2019domain}.
We report Chamfer Distance (CD) and Intersection over Union (IoU) scores for our reconstructions. For CD, we apply Marching Cubes algorithm~\cite{lorensen1987marching} to obtain a point cloud and randomly sample 1024 points from it. For IoU, we downscale our reconstructed shapes to $32^3$ in order to ensure consistency with other reported baselines. Note that we are not able to report IoU scores for the baselines that use point cloud or mesh representations.  Details about the evaluation metrics and reported baselines can be found in the supplementary material.

\begin{table}[t]
\caption{\textbf{3D Shape Reconstruction results on Pix3D chairs.} All \textbf{supervised} methods are trained using paired ShapeNet chair shapes and renderings. We report results of \textbf{our self-supervised} approaches (SIST) for voxel (V) and implicit field (IF) decoder types and unpaired ShapeNet chair shapes with ShapeNet chair renderings (SNR) or Pix3D images (P3D).} 
\label{tab:pix_comparison}
\centering
{{
\begin{tabular}{c| l|c|c}    
Self-supervised & \textbf{Method} &   \textbf{CD} $\downarrow$  &  \textbf{ IoU }$\uparrow$  \\
\hline
\hline
{\color{red} \xmark} &3D-R2N2~\cite{choy20163d} &  $ 0.239  $ & $ 0.136  $  \\
{\color{red} \xmark}& 3D-VAE-GAN~\cite{wu2016learning}  &  $ 0.182  $ & $ 0.171  $\\
{\color{red} \xmark} &  MarrNet~\cite{wu2017marrnet}   &  $  0.144 $ & $ 0.231  $\\
{\color{red} \xmark}& DRC~\cite{tulsiani2017multi} &  $ 0.160  $ & $  0.265 $\\
{\color{red} \xmark}& ShapeHD~\cite{wu2018learning}   &  $  0.123 $ & $ \mathbf{0.284} $ \\
{\color{red} \xmark}& DAREC-vox~\cite{pinheiro2019domain}   &  $ 0.140 $ & $ 0.241 $ \\
{\color{red} \xmark}& DAREC-pc~\cite{pinheiro2019domain}  &  $  \mathbf{0.112} $ & $ - $ \\

\hline
{\color{red} \xmark}& PSGN~\cite{fan2017point} &  $  0.199 $ & $  - $ \\
{\color{red} \xmark}& AtlasNet~\cite{groueix2018atlasnet}   &  $ 0.126  $  & $ -  $ \\

\hline
\hline
{\color{green} \cmark} &  SIST (V+SNR)  &  $ 0.315  $ & $ 0.093  $ \\
{\color{green} \cmark} &  SIST (IF+SNR) &  $ 0.144  $ & $  \mathbf{0.264} $ \\

{\color{green} \cmark} &  SIST (V+P3D)  &  $ \mathbf{0.135}  $ & $ 0.213  $ \\
{\color{green} \cmark} &  SIST (IF+P3D)  &  $ 0.137  $ & $  0.235 $ \\

\end{tabular}}}
\end{table}

\paragraph*{Training Data.}

All of the reported baselines in this experiment use ShapeNet chairs and their renderings for training.
The dataset contains texture data for each synthetic CAD model. So, one can render RGB images from different viewpoints to provide supervision. In our experiments, we use the renderings provided by 3D-R2N2~\cite{choy20163d}. However, for training our networks we assume that we do not know image-shape correspondences and use unpaired data. 

\begin{figure*}[t]
\centering
\includegraphics[width=0.8\textwidth]{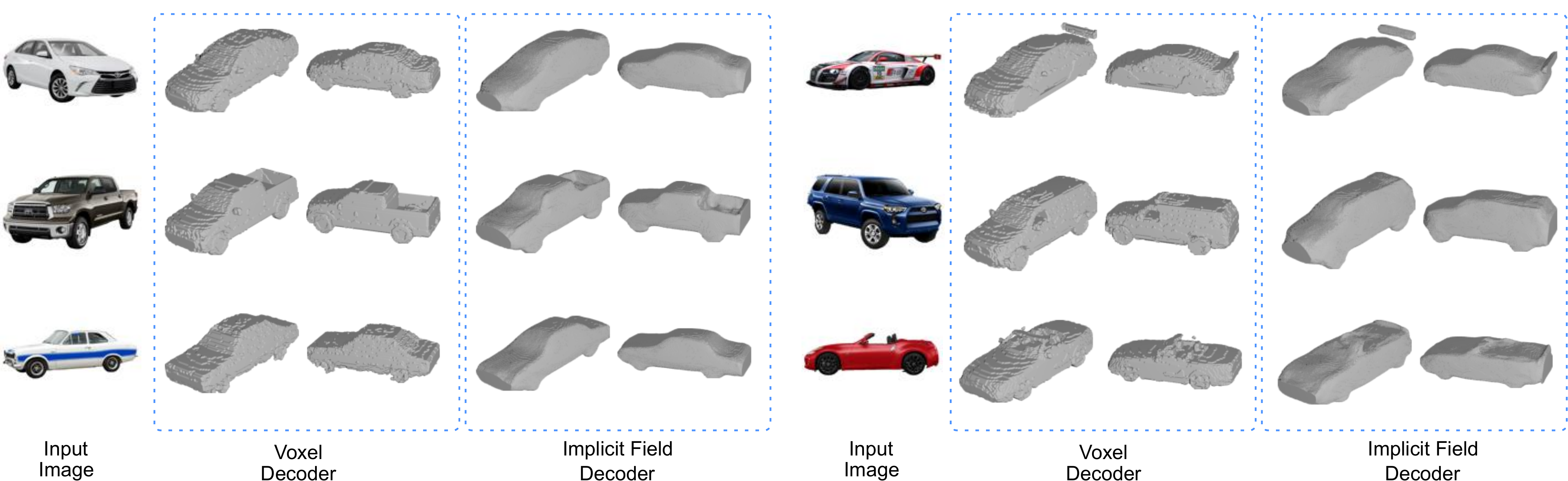}
\caption{\textbf{Single View Shape Reconstruction.} Car images and 3D reconstructions obtained with voxel or implicit field decoders. }
\label{fig:car_svr}
\end{figure*}

 \paragraph*{Pix3D Chairs.}
Table~\ref{tab:pix_comparison} shows 3D shape reconstruction quantitative results on Pix3D chairs obtained by our self-supervised approach in comparison to state-of-the-art supervised methods. We report scores for the voxel and implicit field decoders separately. Our method does not explicitly use strong supervision as other methods require, but still achieves very competitive results. 
Note that our method requires RGB images with clean background for training and testing. 
From other proposed methods, PSGN~\cite{fan2017point} and AtlasNet~\cite{groueix2018atlasnet} also require masks to eliminate the background. Although we compare our results with PSGN~\cite{fan2017point} and AtlasNet~\cite{groueix2018atlasnet}, we also report scores of other baselines that don't use background masks in testing for the sake of completeness. Note that some of these methods use additional data in training.
For example, MarrNet~\cite{wu2017marrnet}, DRC~\cite{tulsiani2017multi} and ShapeHD~\cite{wu2018learning} require groundtruth 2.5D sketches for training. Similarly, DAREC~\cite{pinheiro2019domain} uses natural RGB images for domain adaptation.

\begin{figure}[t]
\centering
\includegraphics[width=0.45\textwidth]{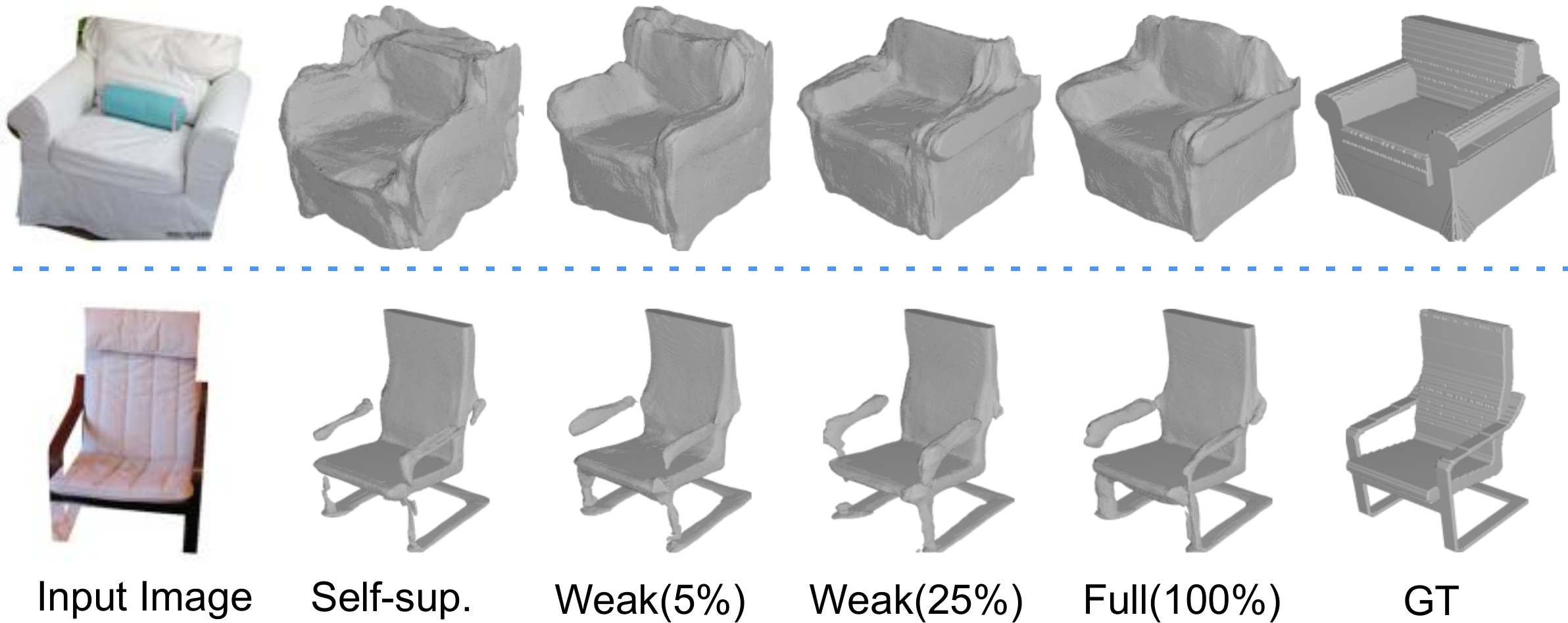}
\caption{ Shape reconstructions on Pix3D chairs for \textbf{different supervision rates} using the implicit field representation (sampled at $512^3$). The reconstruction quality improves with the supervision.  }
\label{fig:chair_svr}
\end{figure}

\paragraph*{Controlling Image Domain.} 
DAREC~\cite{pinheiro2019domain} uses additional images to guide their training. The authors collect chairs from ImageNet and learn latent representations from them. Then, they introduce an adversarial loss that maps ShapeNet renderings and ImageNet images to the same distribution. By doing it, they aim to eliminate the domain adaptation problem. Our method also controls the image domain by using an adversarial loss on generated images. For this reason, we repeat the previous experiment using the Pix3D images instead of ShapeNet renderings. For this task, we use $75\%$ of the Pix3D chair images in training and use the rest in testing. For 3D shapes, we use ShapeNet chairs as before. 

The results of the additional experiment are presented in Table~\ref{tab:pix_comparison}. We are aware of the fact that using a different image set for training is not fair for comparison. Still, we present our results to show that we are not restricted to use a specific image dataset for training. Using a different image dataset can have a huge impact on reconstruction quality. 
Although we obtained comparable scores for the implicit field setting, the performance of the voxel decoder got much better after switching to Pix3D images from ShapeNet renderings. For this reason, we conclude that implicit field setting is more robust and we set the implicit field representation as the default setting of our method.

\subsubsection{Weak and Full Supervision}

In this part, we conduct a comparative study to examine the effect of inserting a small amount of paired data to our training set. In this weakly-supervised setting, we split each iteration into two stages. In the first stage, we update all of the network parameters with the loss function in Equation \eqref{eq:full}. In the second stage, we only train the shape reconstruction network with the paired image-shape examples. Moreover, we train in full supervision to get a reference on the accuracy our method can achieve.

\begin{table}[t!]
\caption{Comparison results on test Pix3D chairs for our method trained with different supervision on Pix3D images and shapes.}
\label{tab:pix_chairs}
\centering
{{
\begin{tabular}{l| c c | c c }

& \multicolumn{2}{c|}{\textbf{Implicit Decoder}}  & \multicolumn{2}{c}{\textbf{Voxel Decoder}} \\
Training  & {CD} $\downarrow$  & { IoU }$\uparrow$  & {CD} $\downarrow$  & { IoU }$\uparrow$ \\
\hline
\hline
 Self-supervised ($0\%$) &  $ 0.095  $ & $ 0.387  $ &  $  0.124 $ & $ 0.421 $ \\
\hline
 Weak Sup.    ($5\%$)   &  $ 0.092  $ & $ 0.395 $ &  $ 0.105 $ & $0.509 $\\
 Weak Sup.  ($25\%$)   &  $ 0.084  $ & $ 0.441 $ &  $ 0.094  $ & $ 0.532 $ \\
 \hline
   Full Sup.  ($100\%$)   &  $ 0.083  $ & $ 0.453 $ &  $  0.104 $ & $ 0.550 $\\
\end{tabular}}}
\end{table}

\begin{figure*}[t]
\centering
\includegraphics[width=0.75\textwidth]{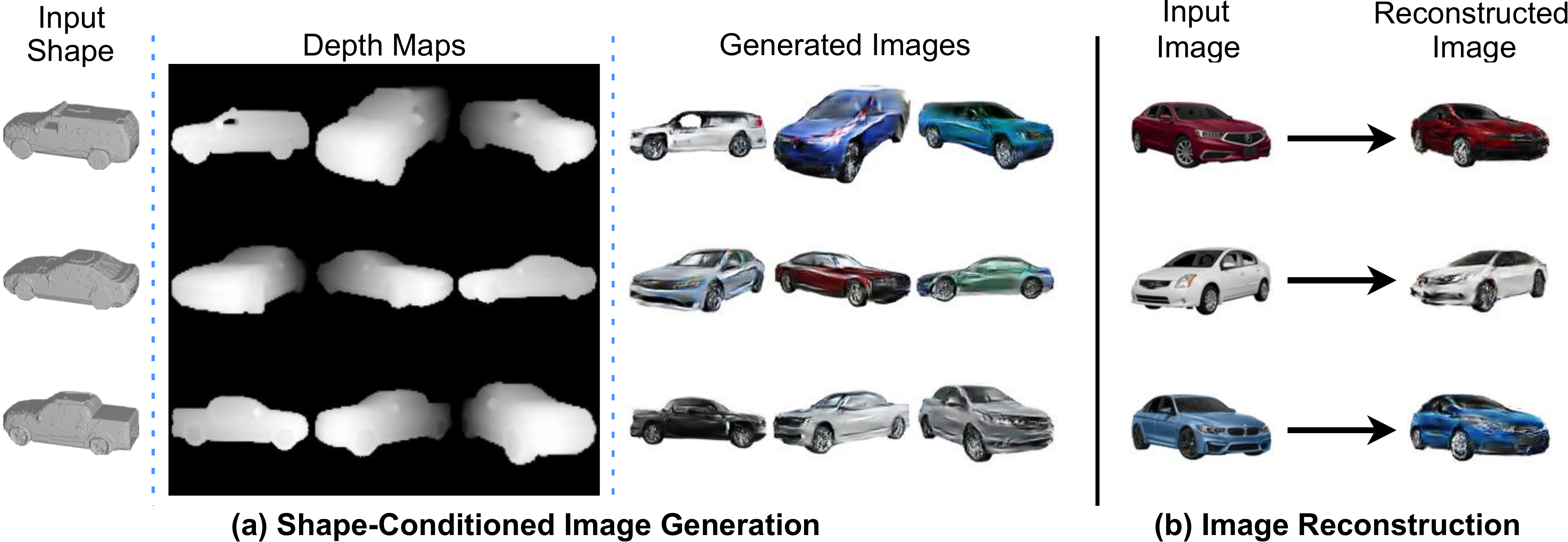}
\caption{\textbf{Shape-Conditioned Image Generation(a).} Generated images from rendered depth maps corresponding to input shapes, with viewpoints and appearance vectors randomly sampled. \textbf{Image Reconstruction(b).} Example car images (left) and their generated reconstructions obtained using the estimated 3D shape, appearance code and viewpoint (right).}
\label{fig:generation_reconstruction}
\end{figure*}

\paragraph*{Pix3D Chairs.} 
We again make use of Pix3D dataset which contains image and shape pairs. In Table~\ref{tab:pix_chairs}, the scores obtained for self-supervised, weakly supervised and full supervised cases are compared for both decoder types. In this experiment, we create our paired data by randomly selecting $5\%$ and $25\%$ of the training images, respectively. Figure~\ref{fig:chair_svr} demonstrates reconstructions achieved using different supervision rates. The results verify that using paired data boosts the performance of our reconstructions. It is also clear that the results tend to get better as the supervision rate increases. 
Moreover, we observed that better CD scores are achieved with implicit field representation while voxels provide better results in IoU scores.      
We also invite the reader to check the supplementary material for the evaluations of weak supervision on multi-class training.

\subsection{Visual Applications}

Our method can be deployed for a large number of tasks involving object image and 3D shape manipulation and translation. In this section, we briefly describe several tasks and present visual results for each. For this purpose, we trained our network with the car dataset of VON~\cite{zhu2018visual} which uses ShapeNet models. 

\noindent\textbf{Single View Shape Reconstruction.}
The primary objective of our approach is to be able to reconstruct the full shape from an example RGB image. In Figure~\ref{fig:car_svr}, we demonstrate the reconstructions obtained using our shape reconstruction network for both decoder types. For voxels, we provide our results using a grid of $128^3$ which is the resolution of the dataset. On the other hand, the implicit field decoder enables us to generate shapes with an arbitrary resolution by adapting the sampling resolution. Therefore, we prefer to provide our reconstructions with $512^3$ sampling rate. 

Both decoder selections result in accurate 3D reconstructions that align with the RGB image. We observed that using implicit fields results in smoother reconstructions as it uses MLPs to generate shapes. Although the inference time required is longer, we don't have any restrictions on the resolution of the reconstructed shapes.
For this reason, we use implicit field representation in the following applications.

\begin{figure*}[t]
\centering
\includegraphics[width=0.8\textwidth]{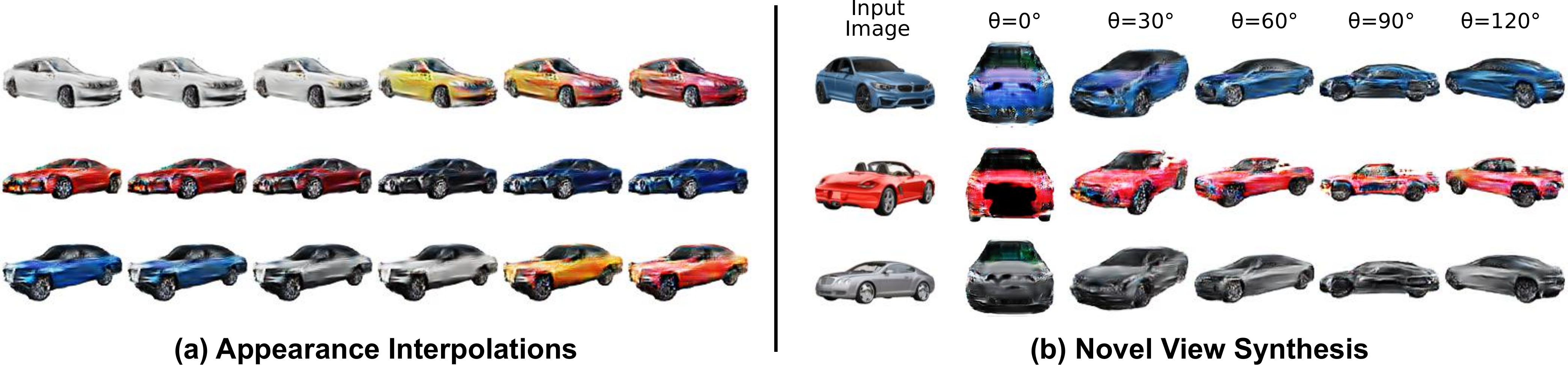}
\caption{\textbf{Appearance Interpolation(a).} Our method can smoothly interpolate between different surface textures while preserving shape and viewpoint. \textbf{Novel View Synthesis(b).} Generated novel views of the example input images from different viewpoints.}
\label{fig:appearance_nvs}
\end{figure*}

\noindent\textbf{Shape-Conditioned Image Generation.}
Our image generation network maps a depth map and an appearance code to a realistic RGB image. in Figure~\ref{fig:generation_reconstruction}, we demonstrate a few images which are generated by sampling an appearance code from a Gaussian distribution.

\noindent\textbf{Image Reconstruction.}
Our method extracts shape, appearance and viewpoint features from an RGB image. Here, we extract these features from an input image and combine them again to generate a reconstruction. Different from other representation learning methods, our method explicitly generates the shape first and then renders a depth map from it to generate the new image. Figure~\ref{fig:generation_reconstruction} shows some of the reconstructed images. Note that the viewpoint must be estimated accurately to reconstruct the input image.

\noindent\textbf{Novel View Synthesis.}
We have demonstrated that we are able to reconstruct the input image from shape, appearance and viewpoint representations. Now, we replace the viewpoint estimation with the angles we want to generate the image. In other words, we generate novel views of the input RGB by setting the viewpoint code with an arbitrary value. In Figure~\ref{fig:appearance_nvs}, we demonstrate the novel views generated by our method. For this experiment, we set the elevation angle to $10^{\circ}$ and modify the azimuth angle to generate images from different viewpoints.

\noindent\textbf{Appearance Interpolation.}
Given a 3D shape, we can generate RGB images by sampling $z_a$ from a Gaussian distribution. Now, we demonstrate that we can interpolate between two randomly sampled appearance codes in the latent space. In Figure~\ref{fig:appearance_nvs}, we show the generated images obtained by these interpolations. In this setting, we used the optimal transport map~\cite{agustsson2017optimal} interpolations instead of interpolating linearly in the latent space.

\begin{figure*}[t]
\centering
\includegraphics[width=0.82\textwidth]{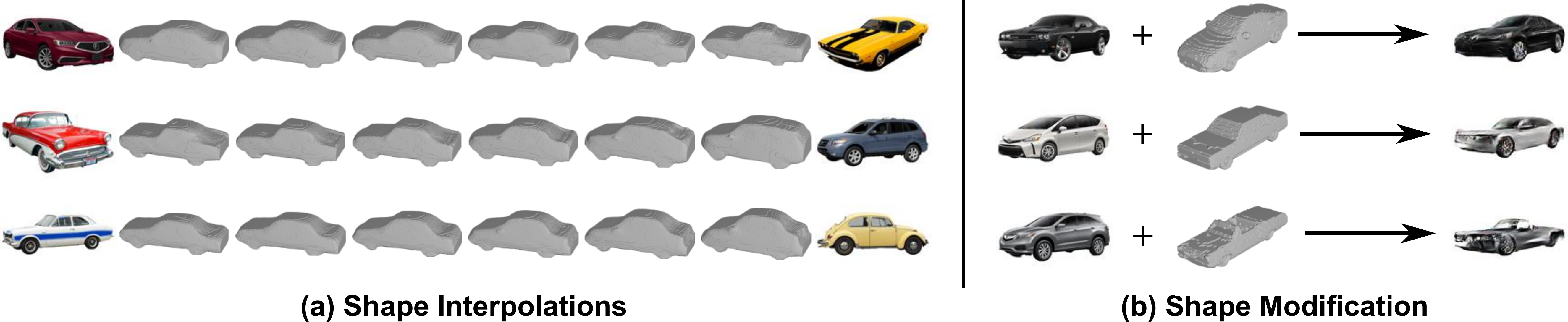}
\caption{\textbf{Shape Interpolation(a).} Our method can smoothly interpolate between different shapes. \textbf{Shape Modification(b).} The shape of an image is modified while keeping appearance and viewpoint the same. }
\label{fig:shape_int_modification}
\end{figure*}

\noindent\textbf{Shape Interpolation.}
Similar to the interpolations we did to appearance vectors, we are able to perform interpolations in the 3D space. In Figure~\ref{fig:shape_int_modification}, we sample and interpolate between two different latent representations and use them as shape codes to generate car shapes. Note that we obtain smooth transitions when interpolating in the latent space. 

\noindent\textbf{Shape Modification.}
Here, we demonstrate that we can infer texture information from an example image and transfer it to a real shape. To do that, we obtain appearance vector $z_a$ and viewpoint $z_v$ from an input image and sample another 3D model from the ShapeNet dataset. Then, we create the depth map and combine it with the inferred appearance code to generate a new RGB image. Examples are shown in Figure~\ref{fig:shape_int_modification}. Note that the generated samples possess the shape characteristics imposed by the 3D model, but exhibit the same appearance and viewpoint characteristics transferred from the input RGB images.

\subsection{Ablation Study}

For the ablation study, we trained our networks with the car dataset provided by VON~\cite{zhu2018visual}. We only modified the appearance $\lambda_A$ and viewpoint $\lambda_V$ terms in Equation~\eqref{eq:cyclic} to observe their effect on image reconstruction and keep other parameters same as stated in implementation details. 

\begin{figure}[t]
\centering
\includegraphics[width=0.35\textwidth]{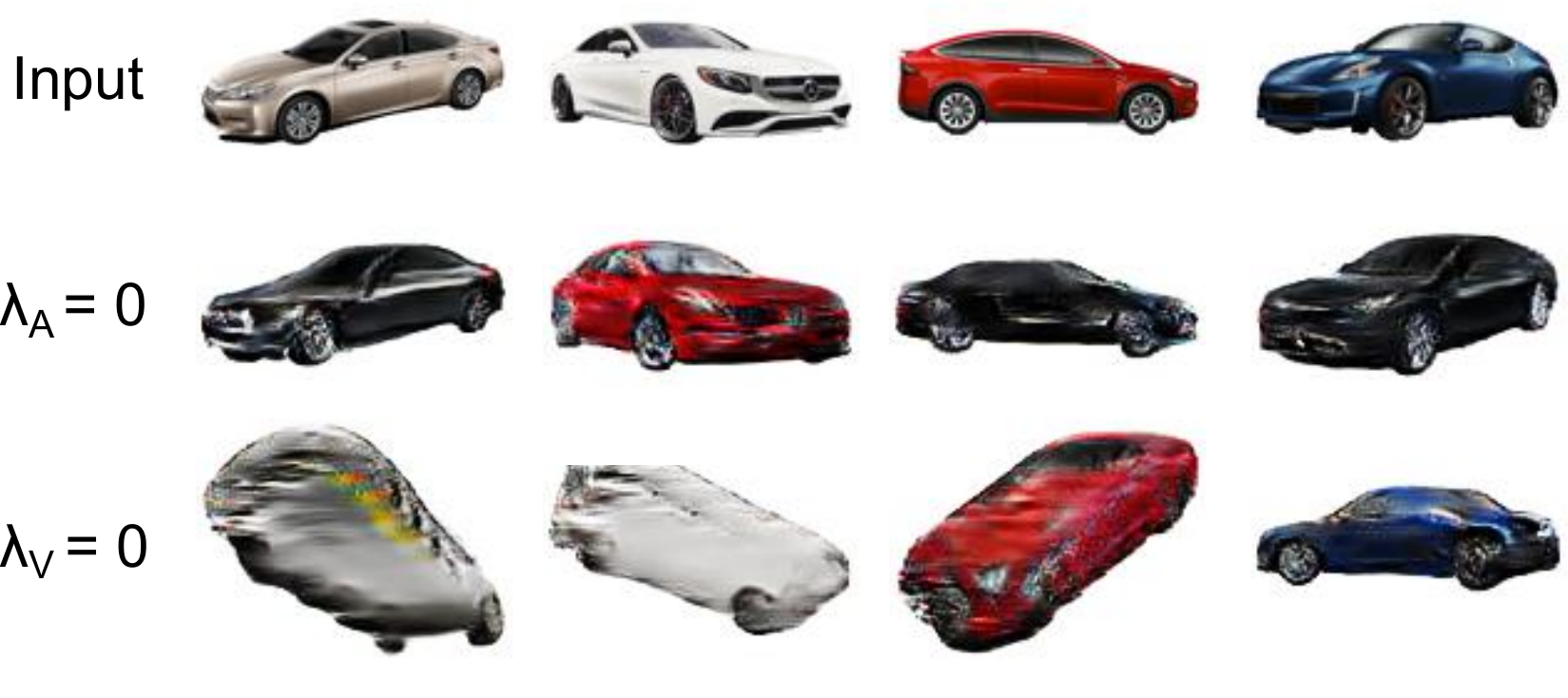}
\caption{\textbf{Ablation.} Example of car images and their reconstructions achieved by using our network trained by setting appearance $\lambda_A$ and viewpoint $\lambda_V$ to 0, respectively, in~\eqref{eq:cyclic}. }
\label{fig:ablation}
\end{figure}

\noindent\textbf{Training without Appearance Loss.} We set the appearance cyclic loss term $\lambda_A$ to zero (no appearance encoder) in \eqref{eq:cyclic} to see its effect on image reconstruction. 
In Figure~\ref{fig:ablation}, we demonstrate input images and reconstructed images using the representations learned by our method. The images show that the shape and pose information is preserved. On the other hand, the texture of the generated sample is determined randomly. 

\noindent\textbf{Training without Viewpoint Loss.} We set $\lambda_V$ to zero in \eqref{eq:cyclic} and eliminate the effect of viewpoint encoder from our framework. Figure~\ref{fig:ablation} shows  input images and the reconstructions achieved by the learned representations. We observe that the method usually fails to generate realistic images as the viewpoint encoder cannot estimate pose angles.

\section{Conclusions}

In this paper, we have proposed SIST, a novel method for translating between RGB images and 3D shape representations trained using unpaired datasets of the same object category. To do it, we first introduced a generative model that generates an image from an object shape. Then, we used the generated image to train our shape reconstruction network. With this self-supervised training method, we obtained results competitive with the state-of-the-art reconstruction methods trained in full supervision. We also proposed a weakly-supervised setting to further improve our shape reconstruction results. 

Our method demonstrated impressive results in learning disentangled features from images. These representations were demonstrated for practical applications such as novel view synthesis and several shape and appearance modifications. Different from other approaches, we were able to use implicit fields in reconstruction in addition to the voxels. In the future, our method can be extended to other representations such as point clouds or meshes.

{\small
\bibliographystyle{ieee}
\bibliography{egbib}
}

\appendix

\begin{centering}
   \twocolumn[\section*{\centering\Large Self-Supervised 2D Image to 3D Shape Translation with Disentangled Representations - Supplementary Material}
      \vspace{0.2cm}
] 

\end{centering}

\section{Datasets}

For training our SIST framework, we use unpaired image and shape datasets as we aim to perform self-supervised training and learn translations between two separate domains. Therefore, we require datasets which have images and shapes of the same object class. Below are the details of the datasets used in our experiments and in the main paper.

\paragraph{ShapeNet~\cite{chang2015shapenet}.}
ShapeNet is a large dataset with 55 different object categories. It contains approximately 51300 unique CAD models of these object classes. It also contains models of 12 object classes from the well-known 3D object repository, PASCAL3D+ \cite{xiang2014beyond}. For \textit{car} and \textit{chair} object classes, ShapeNet contains 3513 and 6777 models respectively. ShapeNet is an information-rich repository which provides data with physical, geometric, texture, and language-related annotations. However, such annotations are not explicitly needed since we only need the shape. 
The full dataset is publicly available at:\\ \url{https://www.shapenet.org/}

\paragraph{Pix3D~\cite{sun2018pix3d}.}
Pix3D is a recently created repository which contains image-shape pairs. It is suitable for many shape related tasks such as single view reconstruction, shape retrieval etc. It is created by extending the IKEA furniture repository~\cite{lim2013parsing} and contains several object classes such as chairs, desks, tables and beds. The full dataset contains 219 shape models and 14600 RGB images which are collected through a web search. We use Pix3D dataset to evaluate or shape reconstruction results which are trained on ShapeNet dataset as our unpaired datasets do not have groundtruth data. 
The Pix3D dataset is accessible at the following link:\\ \url{https://github.com/xingyuansun/pix3d}

\paragraph{VON~\cite{zhu2018visual}.} 
We perform our experiments on cars since we have unpaired RGB images and CAD models of this class presented in VON~\cite{zhu2018visual}. VON uses ShapeNet~\cite{chang2015shapenet} dataset for shapes. For RGB images they create their own dataset by collecting clean background images from Google image search. The dataset contains 2605 car images in total. $75\%$ of the images are used in the training and the others are used in the test. 
VON repository is publicly available:\\ \url{ https://github.com/junyanz/VON } 

\section{Evaluation Metrics}

For single view reconstruction task, we use the Chamfer Distance and the Intersection over Union metrics to evaluate the reconstruction quality. These metrics are standard and commonly employed in the related literature~\cite{pinheiro2019domain,wu2018learning,groueix2018atlasnet,tulsiani2017multi}.

\paragraph{Chamfer Distance (CD).}
 Chamfer Distance is a metric which is used to measure the distance between two point clouds. In our case, we want to use CD to measure the quality of the surface reconstruction.
 For this reason, we first convert our voxelized shape output into mesh structure using marching cubes algorithm~\cite{lorensen1987marching}. Then, we only consider the vertices of the triangular structures and create a point cloud. Similarly, we have the vertices of the CAD data provided with the dataset. 
 
 The value of the CD is dependent on the scale of the data and the number of points sampled when converting voxels into point clouds. In order to be consistent with other baselines, we linearly scale our point cloud such that the longest dimension has unit length. We also randomly sample 1024 points from the point cloud. 
 After applying all these processing steps, we calculate the CD between two point clouds $P_1$ and $P_2$ as follows:
 
 \begin{equation}
 \begin{aligned}
 CD(P_1, P_2) & = &\frac{1}{| P_1 |} \sum_{a \in P_1} \min_{b \in P_2} \left\|a-b \right\|_2 \\
 & + & \frac{1}{| P_2 |} \sum_{a \in P_2} \min_{b \in P_1} \left\|a-b \right\|_2
\end{aligned}
\end{equation}

where $a,b \in \mathbb{R}^3$ are points from the clouds. This expression 
calculates the average of closest point distances between two sets. For this reason, obtaining lower scores means better reconstructions.

\paragraph{Intersection over Union (IoU).}
Intersection over union is a common metric which is generally used for object detection \& segmentation tasks. It is calculated by dividing the overlapped area of two sets $A_1$ and $A_2$ to the union. For voxel data, we downscale our reconstructed shape to $32^3$ in order to be consistent with other reported baselines. 

\begin{equation}
IoU(A_1, A_2) = \frac{A_1 \cap A_2}{A_1 \cup A_2}
\end{equation}

We used the same code provided by Pix3D~\cite{sun2018pix3d} for CD and IoU implementations so that our results are consistent with other reported baselines. The evaluation code can be accessed with the following link: \\ \url{https://github.com/xingyuansun/pix3d}

\section{Reported Baselines}

\noindent\textbf{3D-R2N2}~\cite{choy20163d} uses a recurrent neural network structure to generate a voxel occupancy grid. This method is designed to process a single image or a sequence of images. The reconstruction is refined in each stage as more views of the same object is given. However, this method is evaluated for single images in this paper.

\noindent\textbf{3D-VAE-GAN}~\cite{wu2016learning} is a generative model that maps a probabilistic latent space to 3D voxels. Different from other methods, it uses adversarial training to generate realistic voxel shapes.

\noindent\textbf{MarrNet}~\cite{wu2017marrnet} is a method which uses the 2.5D sketches for shape reconstruction. It first estimates 2.5D sketches with a network and performs 3D shape estimation. It also introduces a reprojection consistency loss such that estimated sketches and 3D shapes are consistent.

\noindent\textbf{DRC}~\cite{tulsiani2017multi} proposes a differentiable ray consistency (DRC) formulation to improve reconstruction quality. It also represents the shape in a probabilistic occupancy grid.

\noindent\textbf{ShapeHD}~\cite{wu2018learning} also estimates 2.5D sketches from input image like MarrNet. Then it reconstructs the shape and uses another adversarial module to enforce naturalness. 

\noindent\textbf{DAREC}~\cite{pinheiro2019domain} also uses an adversarial loss to get better voxel or point cloud reconstructions. However, it applies this loss to the latent space where training images are encoded. In addition to this, DAREC uses another set of unlabeled natural images and tries to map them to the same distribution as other images. So, this method is trained with synthetic renderings but performs well on natural RGB images as well.

\noindent\textbf{PSGN}~\cite{fan2017point} proposes an architecture to train a conditional sampler, which effectively samples points. This generator model creates a point cloud to represent 3D shape which is the reason that we cannot provide IoU scores for this approach.

\noindent\textbf{AtlasNet}~\cite{groueix2018atlasnet} represents a shape with a collection of parametric surface element. Different from other methods, it generates a mesh representation which is not closed. We also cannot report IoU scores because of this reason.

\begin{table}[t]
\caption{\textbf{3D Shape Reconstruction results on Pix3D chairs.} All \textbf{supervised} methods are trained using paired ShapeNet chair shapes and renderings. We report results of our approaches (SIST) for voxel (V) and implicit field (IF) decoder types and ShapeNet chair shapes with ShapeNet chair renderings (SNR) or Pix3D images (P3D).}

\label{tab:pix_comparison2}
\centering
{{
\begin{tabular}{c| l|c|c}    
Self-supervised & \textbf{Method} &   \textbf{CD} $\downarrow$  &  \textbf{ IoU }$\uparrow$  \\
\hline
\hline
{\color{red} \xmark} &3D-R2N2~\cite{choy20163d} &  $ 0.239  $ & $ 0.136  $  \\
{\color{red} \xmark}& 3D-VAE-GAN~\cite{wu2016learning}  &  $ 0.182  $ & $ 0.171  $\\
{\color{red} \xmark} &  MarrNet~\cite{wu2017marrnet}   &  $  0.144 $ & $ 0.231  $\\
{\color{red} \xmark}& DRC~\cite{tulsiani2017multi} &  $ 0.160  $ & $  0.265 $\\
{\color{red} \xmark}& ShapeHD~\cite{wu2018learning}   &  $  0.123 $ & $ \mathbf{0.284} $ \\
{\color{red} \xmark}& DAREC-vox~\cite{pinheiro2019domain}   &  $ 0.140 $ & $ 0.241 $ \\
{\color{red} \xmark}& DAREC-pc~\cite{pinheiro2019domain}  &  $  \mathbf{0.112} $ & $ - $ \\
\hline

{\color{red} \xmark}& PSGN~\cite{fan2017point} &  $  0.199 $ & $  - $ \\
{\color{red} \xmark}& AtlasNet~\cite{groueix2018atlasnet}   &  $ 0.126  $  & $ -  $ \\

\hline
\hline
{\color{red} \xmark} &  SIST (IF+SNR) &  $ \mathbf{0.133}  $ & $  \mathbf{0.264} $ \\

\hline
{\color{green} \cmark} &  SIST (V+SNR)  &  $ 0.315  $ & $ 0.093  $ \\
{\color{green} \cmark} &  SIST (IF+SNR) &  $ 0.144  $ & $  \mathbf{0.264} $ \\
{\color{green} \cmark} &  SIST (V+P3D)  &  $ \mathbf{0.135}  $ & $ 0.213  $ \\
{\color{green} \cmark} &  SIST (IF+P3D)  &  $ 0.137  $ & $  0.235 $ \\

\end{tabular}}}
\end{table}

\begin{table*}[t]
\caption{Comparison results on test Pix3D dataset for our method \textbf{(voxel decoder)}
trained with different supervision on Pix3D images and shapes from \textbf{all categories}.}
\label{tab:pix_allcategories_voxel}
\centering
{\resizebox{0.8\linewidth}{!}{
\begin{tabular}{ l | c || c | c | c |c | c | c | c | c | c || c } 

 &  & bed & bkcs & chair & desk & sofa & table & tool & wrdr & misc & \textbf{Avg.} \\

\hline
 \multirow{2}{*}{Self-supervised} & CD $\downarrow$  &  $ 0.232 $   & $  0.176 $  &  $0.237 $  & $ 0.216  $  &  $ 0.145 $  & $ 0.222 $  &  $ 0.076 $   & $ 0.208 $  &  $0.319 $ &  $ 0.211 $\\
 
 & IoU $\uparrow$  &  $ 0.201 $   & $  0.120 $  &  $0.196 $  & $  0.206 $  &  $ 0.444 $  & $ 0.186 $  &  $ 0.308 $   & $ 0.145 $  &  $ 0.053$ &  $  0.247$\\
 \hline
 
  \multirow{2}{*}{Weak-Sup.($5\%$)}& CD $\downarrow$  &  $ 0.140 $   & $ 0.111  $  &  $0.173 $  & $ 0.148  $  &  $0.093  $  & $ 0.173 $  &  $ 0.134 $   & $ 0.138 $  &  $ 0.271 $ &  $ 0.151 $\\
  
   & IoU $\uparrow$  &    $ 0.416 $   & $ 0.468  $  &  $ 0.328$  & $ 0.363 $  &  $  0.686$  & $ 0.363 $  &  $ 0.257 $   & $ 0.578 $  &  $ 0.105 $ &  $ 0.421 $\\
 \hline
 
   \multirow{2}{*}{Weak-Sup.($25\%$)}& CD $\downarrow$  &  $ 0.095 $   & $  0.092 $  &  $0.138 $  & $  0.115 $  &  $ 0.075 $  & $ 0.134 $  &  $ 0.109 $   & $ 0.097 $  &  $ 0.238$ &  $ 0.118 $\\
  
   & IoU $\uparrow$  &  $ 0.614 $   & $ 0.501  $  &  $ 0.403$  & $  0.474 $  &  $0.775  $  & $ 0.446 $  &  $ 0.264 $   & $ 0.747 $  &  $0.195 $ &  $ 0.516 $\\
   
 \hline

\multirow{2}{*}{Full-Sup.($100\%$)}& CD $\downarrow$  &  $ 0.078 $   & $  0.073 $  &  $ 0.174$  & $  0.097 $  &  $ 0.062 $  & $ 0.105 $  &  $ 0.137 $   & $ 0.064 $  &  $0.209 $ &  $ 0.119 $\\
  
& IoU $\uparrow$  &  $ 0.714 $   & $  0.592 $  &  $0.380 $  & $ 0.594  $  &  $  0.839 $ & $ 0.567 $  &  $ 0.283 $   & $ 0.865 $  &  $0.238 $ &  $ 0.568 $\\
   
 \hline

\end{tabular}}}
\end{table*}

\begin{table*}[t]
\caption{
Comparison results on test Pix3D dataset for our method \textbf{(implicit field decoder)}
trained with different supervision on Pix3D images and shapes from \textbf{all categories}. }
\label{tab:pix_allcategories_implicit}
\centering
{\resizebox{0.8\linewidth}{!}{
\begin{tabular}{ l | c || c | c | c |c | c | c | c | c | c || c } 

 &  & bed & bkcs & chair & desk & sofa & table & tool & wrdr & misc & \textbf{Avg.} \\

\hline
 \multirow{2}{*}{Self-supervised} & CD $\downarrow$  &  $ 0.193 $   & $ 0.186  $  &  $ 0.142 $  & $ 0.171  $  &  $  0.140$  & $ 0.189  $  &  $ 0.176 $   & $  0.180 $  &  $ 0.253 $ &  $ 0.160 $\\
 
 & IoU $\uparrow$  &  $ 0.260 $   & $  0.199 $  &  $ 0.270 $  & $ 0.220  $  &  $ 0.466 $  & $  0.180 $  &  $  0.200$   & $  0.394 $  &  $ 0.138 $ &  $ 0.289 $\\
 \hline
  \multirow{2}{*}{Weak-Sup.($5\%$)}& CD $\downarrow$  &  $0.137 $  & $ 0.158  $  &  $ 0.117 $  & $ 0.137  $  &  $  0.109 $  & $  0.166 $  &  $ 0.149 $  & $  0.134 $  &  $ 0.250 $ &  $ 0.131 $ \\
  
   & IoU $\uparrow$  &  $ 0.377 $   & $ 0.246  $  &  $ 0.329 $  & $  0.305 $  &  $0.581  $  & $0.225  $  &  $ 0.261 $   & $  0.580 $  &  $ 0.110 $ &  $ 0.365 $\\
 \hline
 
   \multirow{2}{*}{Weak-Sup.($25\%$)}& CD $\downarrow$  &  $0.108 $  & $ 0.110  $  &  $ 0.096$  & $0.123  $  &  $0.081  $  & $ 0.136  $  &  $ 0.149 $  & $ 0.089  $  &  $ 0.180 $ &  $ 0.105 $ \\
  
   & IoU $\uparrow$  &  $0.469 $   & $ 0.335  $  &  $ 0.385 $  & $  0.326 $  &  $ 0.720 $  & $ 0.292 $  &  $ 0.261 $   & $ 0.755  $  &  $ 0.234 $ &  $ 0.439 $\\
 \hline

\multirow{2}{*}{Full-Sup.($100\%$)}& CD $\downarrow$  &  $ 0.096 $   & $  0.102 $  &  $ 0.093$  & $ 0.113  $  &  $ 0.077 $  & $ 0.122 $  &  $0.107 $   & $ 0.080 $  &  $ 0.173$ &  $ 0.098 $\\
  
& IoU $\uparrow$ &  $ 0.530 $   & $  0.354 $  &  $ 0.403$  & $  0.337 $  &  $ 0.727 $  & $ 0.293 $  &  $ 0.278$   & $  0.791 $  &  $0.212 $ &  $  0.457$\\
   
 \hline
 
\end{tabular}}}
\end{table*}

\begin{figure*}[t]
\centering
\includegraphics[width=0.95\textwidth]{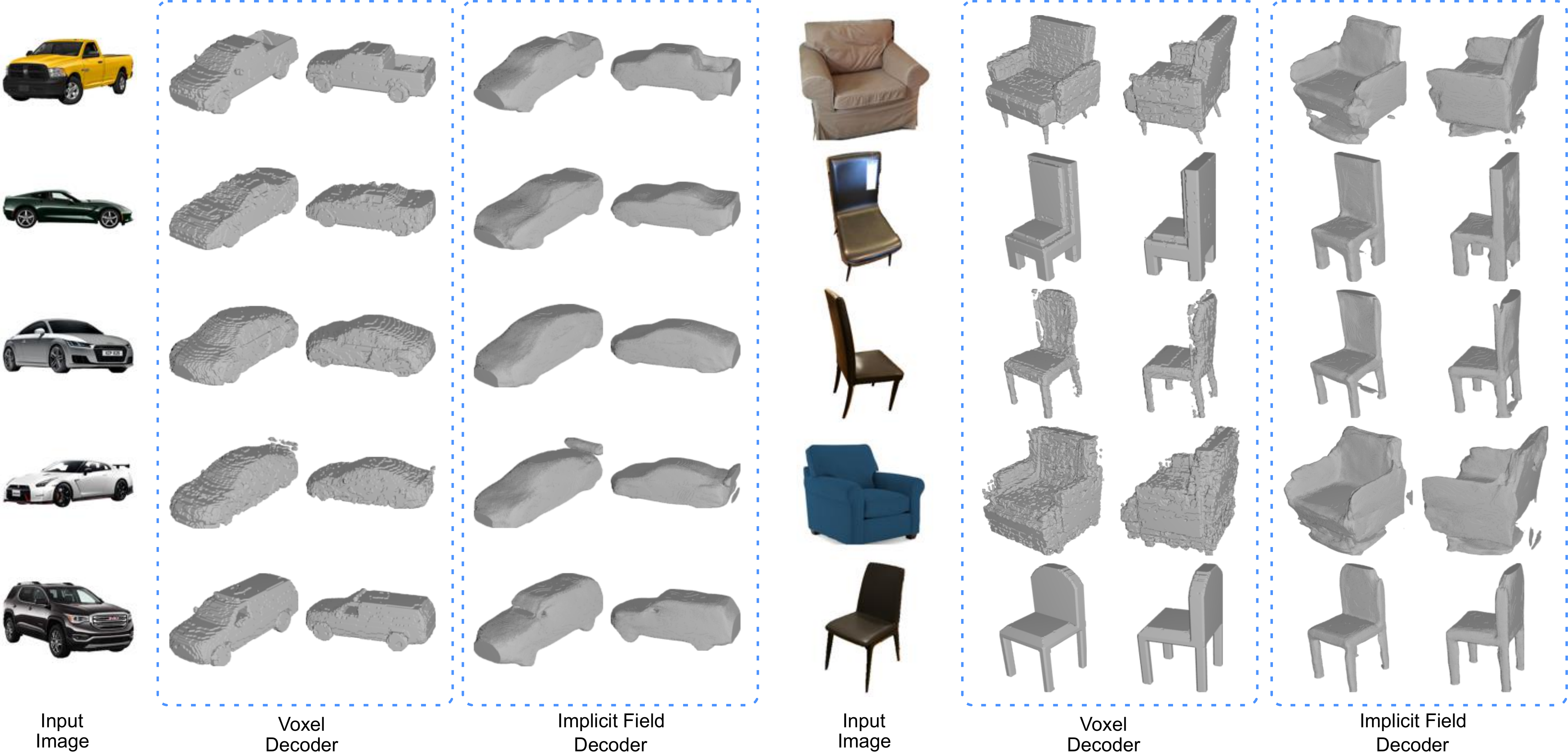}
\caption{\textbf{Single View Shape Reconstruction.} Example RGB images and 3D reconstructions obtained with voxel or implicit field decoders. Different networks are trained for cars and chairs. }
\label{fig:more_examples}
\end{figure*}

\section{Self-supervised vs. Supervised}

Our proposed SIST is a method that relates unpaired image and shape datasets. Table 1 in the main paper compares our self-supervised method with state-of-the-art supervised methods. The reported baselines used ShapeNet chair images and renderings to train their network parameters in full supervision. In this section, we train our default network setting which uses implicit field decoder with full supervision to see where our reconstruction network stands compared to other baselines. For this experiment, we used paired ShapeNet chair images and chairs. Note that we trained our networks for 50 epochs for both supervised and self-supervised settings. 

The results of this additional experiment are presented in Table~\ref{tab:pix_comparison2}. 
The numbers indicate our reconstruction network performs comparably with the other reported baselines. It is also clear that the supervised is better than the self-supervised setting although the results are still comparable.

\section{Weak Supervision on All Categories}

Table 2 in the main paper presents the reconstruction scores for our networks trained with different supervision rates using Pix3D chairs. The results show that using labeled data improves performance. 
Here, we repeat the same experiment using the networks trained with Pix3D images and shapes from all categories. We also evaluate the network for all categories independently. 

Tables~\ref{tab:pix_allcategories_voxel} and~\ref{tab:pix_allcategories_implicit} report the numerical results obtained for single view reconstruction using voxel and implicit field decoders, respectively.
The results verify that weak supervision improves the reconstructions for almost all of the categories.
We observed that the scores are impressive for objects with simple surface typologies and low in-class variation such as sofas and wardrobes. On the other hand, we obtained poorer performance for classes with more complex shapes such as table, tool and misc. Note that tool and misc categories contain a few samples and these samples are totally distinct from each other.

\section{More Examples on Singe View Reconstruction}

In this section, we provide additional visual results for single view reconstruction task. In this setting, we again use VON~\cite{zhu2018visual} dataset. We trained separate networks for car and chair categories. In Figure \ref{fig:more_examples}, we demonstrate the reconstructions achieved using our method trained with cars and chairs.

\section{Failure Cases}

\begin{figure*}[t]
\centering
\includegraphics[width=0.95\textwidth]{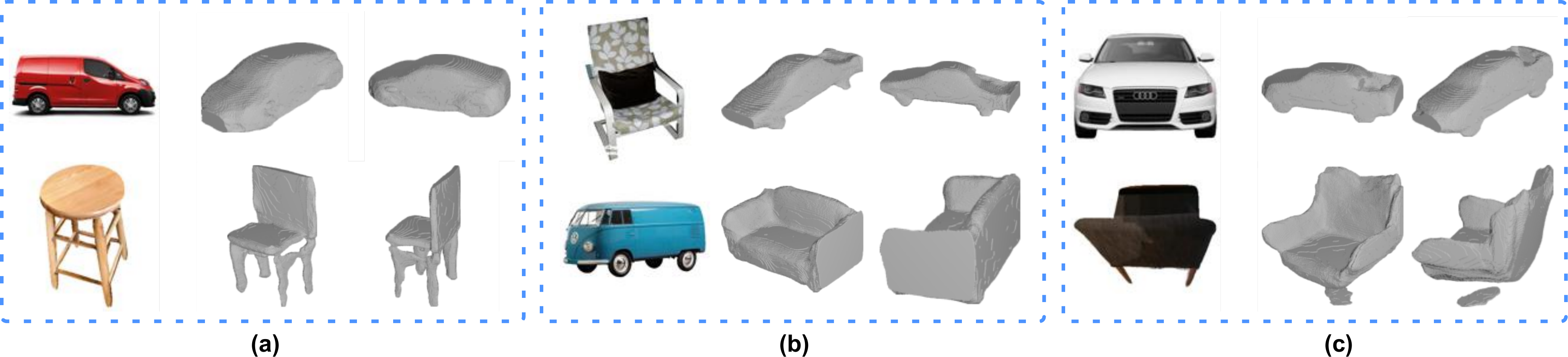}
\caption{ Failure examples resulting from the inconsistency between image and shape datasets (a), out-of-class inputs (b) and the lack of information in the input image (c). We demonstrate failure cases for the networks trained with cars (top) and chairs (bottom) separately.}
\label{fig:failures}
\end{figure*}

In this section, we discuss the limitations of our method by demonstrating some failure cases in 3D reconstruction network. For this purpose, we use distinct networks trained with VON~\cite{zhu2018visual} cars and chairs using the implicit field decoder.  As discussed in \cite{tatarchenko2019single}, CNN based single view reconstruction methods lack per-pixel reasoning ability. Therefore, they perform a recognition task and try to generate the best shape possible from a single feature vector. Due to the fact that the decoder is trained with the models in the training dataset, the reconstruction networks usually fail to generate shapes from unseen classes. In Figure~\ref{fig:failures}, we point to this issue with examples from car and chair classes. Since we are using unpaired datasets, our RGB and shape datasets contain samples which don't match. For example, our RGB dataset contains vehicles like vans, but our training shapes do not. Therefore, the network fails to reconstruct a van, which is an unseen object for it. Similarly, the network trained with chairs cannot reconstruct a stool. 
If we take it one step further and provide images from completely different classes, the network again generates generic cars and chairs.

Another type of reconstruction failure stems from the lack of information in RGB images. In Figure~\ref{fig:failures}, we observe a sedan car is reconstructed as a pickup as the network cannot infer the car type from the front view. Similarly, a chair image is reconstructed poorly because the image does not provide useful cues from that viewpoint.

\section{Network Architectures}

In Tables~\ref{tab:G_image}-\ref{tab:G_shape}, we provide the details for the network architectures of our SIST approach.

\begin{table*}[]
\caption{Architecture of image generator $G_I$}
\label{tab:G_image}
\centering
{
\resizebox{0.6\linewidth}{!}{
\begin{tabular}{c|c|c|c|c|c}    

  \hline
  \textbf{Layer}  &  Output Size & Kernel Size & Stride & BatchNorm & Activation\\
  \hline
  Input:  [depth]  & $128\times 128\times 1$ & & & &\\
  
 [$z_a$] + Conv  & $128\times 128\times 64$ & $7\times7$ &  $1$  & Yes & ReLU \\
 
 [$z_a$] + Conv & $64\times 64\times 128$ & $4\times4$ &  $2$  & Yes & ReLU \\

 [$z_a$] + Conv & $32\times 32\times 256$ & $4\times4$ &  $2$  & Yes & ReLU \\

 [$z_a$] + Conv & $16\times 16\times 512$ & $4\times4$ &  $2$  & Yes & ReLU \\

 Ups($\times2$) + [$z_a$] + Conv & $ 32\times 32\times 256 $ & $ 5\times5 $ &  $1$  & Yes & ReLU \\
 
 Ups($\times2$) + [$z_a$] + Conv & $ 64\times 64\times 128 $ & $ 5\times5 $ &  $1$  & Yes & ReLU \\

 Ups($\times2$) + [$z_a$] + Conv & $ 128\times 128\times 64 $ & $ 5\times5 $ &  $1$  & Yes & ReLU \\
 
  [$z_a$] + [depth] + Conv & $ 128\times 128\times 3 $ & $ 7\times7 $ &  $1$  & No & Tanh \\
\hline
\end{tabular}
}
}
\end{table*}

\begin{table*}[]
\caption{Architecture of image discriminator $D_I$}
\label{tab:D_image}
\centering
{
\resizebox{0.6\linewidth}{!}{
\begin{tabular}{c|c|c|c|c|c}    

  \hline
  \textbf{Layer}  &  Output Size & Kernel Size & Stride & BatchNorm & Activation\\
  \hline
  Input:  [RGB]  & $128\times 128\times 3$ & & & &\\
  
  Conv  & $ 64\times 64\times 64 $ & $ 4\times4 $ &  $2$  & No & LeakyReLU \\
   Conv  & $ 32\times 32\times 128 $ & $ 4\times4 $ &  $2$  & No & LeakyReLU \\
  Conv  & $ 16\times 16\times 256 $ & $ 4\times4 $ &  $2$  & No & LeakyReLU \\
  Conv  & $ 16\times 16\times 512 $ & $ 4\times4 $ &  $1$  & No & LeakyReLU \\
  Conv  & $ 16\times 16\times 1 $ & $ 4\times4 $ &  $1$  & No &  \\
\hline
\end{tabular}
}
}
\end{table*}

\begin{table*}[]
\caption{Architecture of viewpoint encoder $E_V$}
\label{tab:E_view}
\centering
{
\resizebox{0.6\linewidth}{!}{
\begin{tabular}{c|c|c|c|c|c}    

  \hline
  \textbf{Layer}  &  Output Size & Kernel Size & Stride & BatchNorm & Activation\\
  \hline
  Input:  [RGB]  & $128\times 128\times 3$ & & & &\\
  
  Conv  & $ 64\times 64\times 32 $ & $ 4\times4 $ &  $2$  & Yes & ReLU \\
  Conv  & $ 32\times 32\times 64 $ & $ 4\times4 $ &  $2$  & Yes & ReLU \\
  Conv  & $ 16\times 16\times 128 $ & $ 4\times4 $ &  $2$  & Yes & ReLU \\

  Conv  & $ 8\times 8\times 256 $ & $ 4\times4 $ &  $2$  & Yes & ReLU \\

  Conv  & $ 4\times 4\times 512 $ & $ 4\times4 $ &  $2$  & Yes & ReLU \\

  Conv  & $ 1\times 1\times 2 $ & $ 4\times4 $ &  $1$  & No & Tanh \\

\hline
\end{tabular}
}
}
\end{table*}

\begin{table*}[]
\caption{Architecture of appearance encoder $E_A$}
\label{tab:E_texture}
\centering
{
\resizebox{0.6\linewidth}{!}{
\begin{tabular}{c|c|c|c|c|c}    

  \hline
  \textbf{Layer}  &  Output Size & Kernel Size & Stride & BatchNorm & Activation\\
  \hline
  Input:  [RGB]  & $128\times 128\times 3$ & & & &\\
  
  Conv  & $ 64\times 64\times 32 $ & $ 4\times4 $ &  $2$  & Yes & ReLU \\
  Conv  & $ 32\times 32\times 64 $ & $ 4\times4 $ &  $2$  & Yes & ReLU \\
  Conv  & $ 16\times 16\times 128 $ & $ 4\times4 $ &  $2$  & Yes & ReLU \\

  Conv  & $ 8\times 8\times 256 $ & $ 4\times4 $ &  $2$  & Yes & ReLU \\

  Conv  & $ 4\times 4\times 512 $ & $ 4\times4 $ &  $2$  & Yes & ReLU \\

  Conv  & $ 1\times 1\times 512 $ & $ 4\times4 $ &  $1$  & No & ReLU \\
  FCL & $2 \times 16$ & & & &  \\

\hline
\end{tabular}
}
}
\end{table*}

\begin{table*}[]
\caption{Architecture of shape encoder $E_S$}
\label{tab:E_shape}
\centering
{
\resizebox{0.6\linewidth}{!}{
\begin{tabular}{c|c|c|c|c|c}    

  \hline
  \textbf{Layer}  &  Output Size & Kernel Size & Stride & BatchNorm & Activation\\
  \hline
  Input:  [RGB]  & $128\times 128\times 3$ & & & &\\
  
  Conv  & $ 64\times 64\times 32 $ & $ 4\times4 $ &  $2$  & Yes & ReLU \\
  Conv  & $ 32\times 32\times 64 $ & $ 4\times4 $ &  $2$  & Yes & ReLU \\
  Conv  & $ 16\times 16\times 128 $ & $ 4\times4 $ &  $2$  & Yes & ReLU \\

  Conv  & $ 8\times 8\times 256 $ & $ 4\times4 $ &  $2$  & Yes & ReLU \\

  Conv  & $ 4\times 4\times 512 $ & $ 4\times4 $ &  $2$  & Yes & ReLU \\

  Conv  & $ 1\times 1\times 512 $ & $ 4\times4 $ &  $1$  & No & ReLU \\
  FCL & $2 \times 128$ & & & &  \\

\hline
\end{tabular}
}
}
\end{table*}

\begin{table*}[]
\caption{Architecture of voxel decoder. UpConv3d refers to the transposed convolution operation using 3d kernels. }
\label{tab:G_shape}
\centering
{
\resizebox{0.6\linewidth}{!}{
\begin{tabular}{c|c|c|c|c|c}    

  \hline
  \textbf{Layer}  &  Output Size & Kernel Size & Stride & BatchNorm & Activation\\
  \hline
  Input:  [$z_s$]  & $ 128\times 1\times 1 \times 1 $ & & & &\\
  
  UpConv3d  & $ 512\times 4\times 4 \times 4 $  & $ 4\times4\times4 $ &  $1$  & No & ReLU \\
  UpConv3d  & $ 256\times 8\times 8 \times 8 $  & $ 4\times4\times4 $ &  $2$  & Yes & ReLU \\
  UpConv3d  & $ 128\times 16\times 16 \times 16 $  & $ 4\times4\times4 $ &  $2$  & Yes & ReLU \\
  UpConv3d  & $ 64\times 32\times 32 \times 32 $  & $ 4\times4\times4 $ &  $2$  & Yes & ReLU \\
  UpConv3d  & $ 32\times 64\times 64 \times 64 $  & $ 4\times4\times4 $ &  $2$  & Yes & ReLU \\
  UpConv3d  & $ 1\times 128\times 128 \times 128 $  & $ 4\times4\times4 $ &  $2$  & No & Sigmoid \\

\hline
\end{tabular}
}
}
\end{table*}



\end{document}